\documentclass[10pt,twocolumn,letterpaper]{article}

\usepackage{iccv}
\usepackage{times}
\usepackage{epsfig}
\usepackage{graphicx}
\usepackage{amsmath}
\usepackage{amssymb}
\usepackage{adjustbox}

\usepackage[pagebackref=true,breaklinks=true,letterpaper=true,colorlinks,bookmarks=false]{hyperref}

\iccvfinalcopy 

\def\iccvPaperID{10594} 

\ificcvfinal\fi

\newcommand{\hc}[1]{\textcolor{cyan}{[#1]}}

\usepackage{tabularx}
\usepackage{multirow}
\usepackage{booktabs} 

\begin{document}

\title{LSD-StructureNet: Modeling Levels of Structural Detail in 3D Part Hierarchies}

\author{Dominic Roberts$^{1}$ \quad Ara Danielyan$^2$ \quad Hang Chu$^2$ \quad Mani Golparvar-Fard$^1$ \quad David Forsyth$^1$\\
$^1$University of Illinois at Urbana-Champaign \quad $^2$Autodesk AI Lab\\
{\tt\small \{djrbrts2,mgolpar,daf\}@illinois.edu \quad \{ara.danielyan,hang.chu\}@autodesk.com}
}

\maketitle
\thispagestyle{empty}

\begin{abstract}  

Generative models for 3D shapes represented by hierarchies of parts can generate realistic and diverse sets of outputs. However, existing models suffer from the key practical limitation of modelling shapes holistically and thus cannot perform conditional sampling, i.e. they are not able to generate variants on individual parts of generated shapes without modifying the rest of the shape. This is limiting for applications such as 3D CAD design that involve adjusting created shapes at multiple levels of detail. To address this, we introduce LSD-StructureNet, an augmentation to the StructureNet architecture that enables re-generation of parts situated at arbitrary positions in the hierarchies of its outputs. We achieve this by learning individual, probabilistic conditional decoders for each hierarchy depth. We evaluate LSD-StructureNet on the PartNet dataset, the largest dataset of 3D shapes represented by hierarchies of parts. Our results show that contrarily to existing methods, LSD-StructureNet can perform conditional sampling without impacting inference speed or the realism and diversity of its outputs.

\end{abstract}

\section{Introduction}




\begin{figure}[t]
\begin{center}
\includegraphics[width=1.0\linewidth]{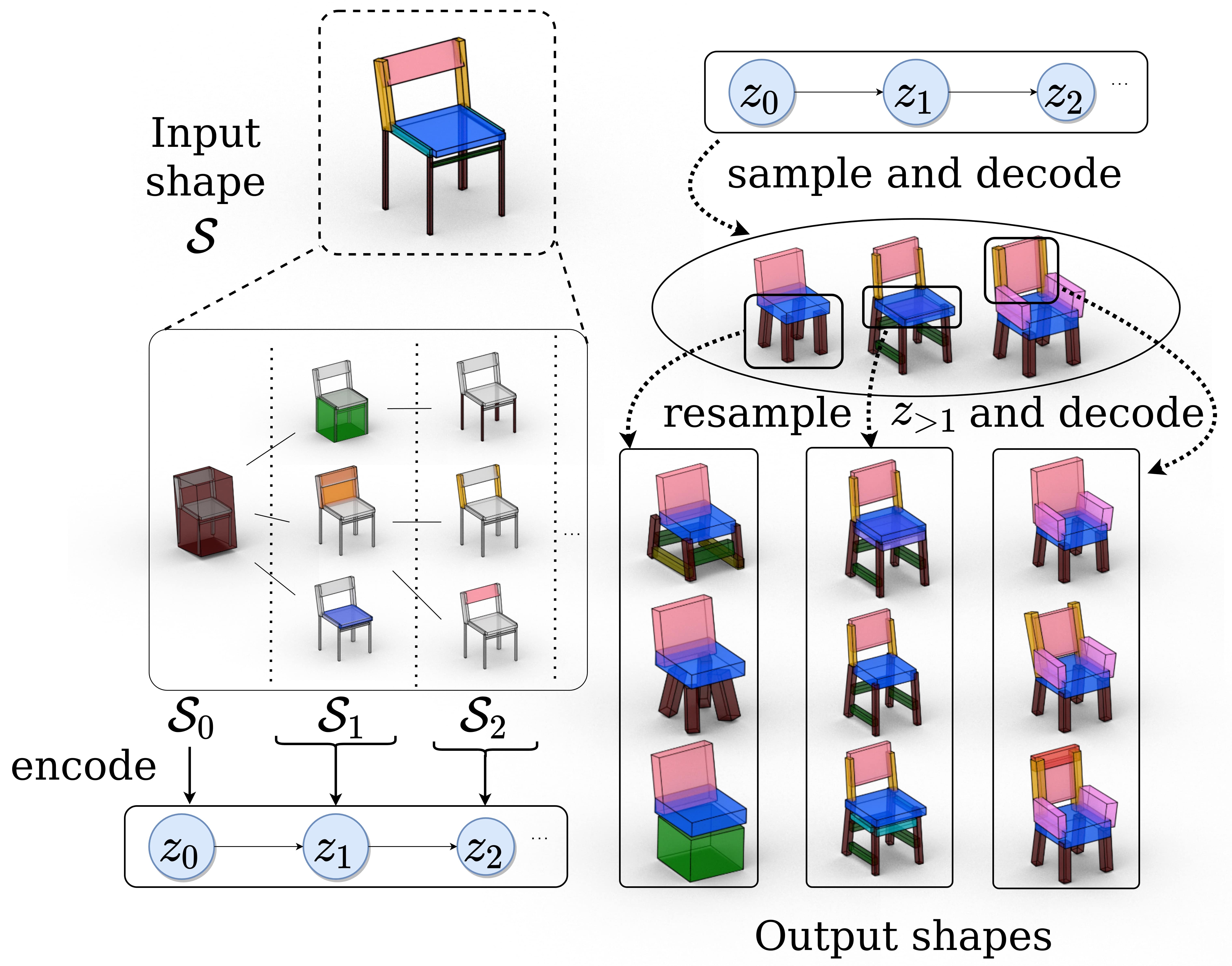}
\end{center}
   \caption{We augment existing methods for generative modelling of 3D shapes structured according to hierarchies so each structural level of detail is modelled with its own generative process. This enables us to generate variants on any sub-tree of the output hierarchy by simply re-sampling and decoding from a sub-sequence of the learnt latent spaces and conditionally on the root node of the sub-tree (e.g. corresponding to the base, seat or back of the chair), without modifying the remainder of the output hierarchy.}
\label{fig:long}
\label{fig:onecol}
\end{figure}

Computer-Aided Design (CAD) involves creating the structure and geometry of 3D objects organized according to hierarchies of parts. This process can be emulated by sampling from generative models trained on datasets of hierarchically structured objects. While existing models  are able to produce novel, diverse and realistic outputs, they only model hierarchies holistically, via single latent spaces that govern the geometry and/or structure of the entirety of their outputs. This makes them ill-suited for use in CAD, as creating objects involves making decisions at multiple levels of detail. For instance, designing a chair consists of both high-level design choices such as deciding whether a chair should have arm-rests or not and low-level design choices such as deciding on the geometry and positioning of the arm-rests.

In this paper, we address this issue by introducing LSD-StructureNet, an augmented version of the StructureNet ~\cite{mo2019structurenet} architecture that can efficiently re-generate arbitrary sub-hierarchies of hierarchically structured 3D objects. Our solution consists of learning one latent space for each depth of an output hierarchy. We term the structure and geometry of elements at the same depth \emph{LSD} (Level of Structural Detail), and generate these elements by sampling from the corresponding latent space and decoding the sample conditionally on the element's parent. We can thus regenerate multiple varying sub-hierarchies of an object without modifying the remainder of the hierarchy by simply re-sampling from the latent spaces at and beyond the appropriate hierarchy depths. In contrast, re-generating a sub-hierarchy with existing generative models cannot be achieved without re-generating the entire hierarchy from scratch and without any guarantee the remainder of the re-generated structure will be at all faithful to the remainder of the original structure.
  
Sets of sub-hierarchies generated using our strategy should be both realistic and diverse in terms of semantics, structure and geometry. This presents a challenge, as data samples in publicly available datasets of hierarchically structured 3D objects are very different from one another and thus do not contain the wide conditional distributions of child parts we seek to learn.  We show that LSD-StructureNet overcomes this challenge, producing outputs that are more realistic and diverse than existing methods. Furthermore, LSD-StructureNet is capable of producing variations on its outputs that differ only beyond a certain LSD. We also show that such outputs are similarly superior to the closest obtainable approximations from StructureNet, and can be obtained far more speedily. 

\section{Related work}

 
\subsection{Unstructured shape generation} 

Generative models can produce realistic shapes in the form of holistic point clouds \cite{qi2017pointnet,shu20193d,valsesia2018learning}; mesh models (e.g. PolyGen~\cite{nash2020polygen} and TM-Net~\cite{gao2020tmnet}); voxels (e.g. SAGNet~\cite{wu2019sagnet,wu2016learning}); octrees~\cite{tatarchenko2017octree}; 3D surfaces~\cite{groueix2018papier}; aggregates of smaller, proxy shapes ~\cite{gadelha2020learning}; and signed distance functions~\cite{park2019deepsdf}. 

In contrast to these models, our shape model
has a hierarchical structure that allows conditional sampling,
which is a common need in applications such as design where objects are organized in 
standardized structures.

\subsection{Structured shape generation} 

Hierarchies appear in recent work, including shape programs
(structured representations from unstructured shapes in ~\cite{tian2018learning}; assembling parts into a whole in ~\cite{zhu2018scores}). ~\cite{kalogerakis2012probabilistic, nash2017shape,schor2019CompoNet} introduced frameworks
that assigned individual probability distributions to
parts of shapes following flattened hierarchies.  Similarly, the latent codes learnt in ~\cite{dubrovina2019composite} were factored according to individual components of shapes, and ~\cite{Wu_2020_CVPR} introduced a Seq2Seq model for flattened object hierarchies. 

In contrast, much recent work has focused on modelling 3D shapes whose parts are structured according to  n-ary tree hierarchies of varying depth ~\cite{mo2019PartNet,li17grass}. Such methods are often built around recursive neural networks ~\cite{socher2011parsing}. StructureNet  ~\cite{mo2019structurenet} and StructEdit ~\cite{mo2020structedit} learn to generate and edit such hierarchical representations respectively. ~\cite{jones2020shapeAssembly} proposed a VAE model for generating assembly programs capable of generating hierarchically structured 3D shapes upon execution; but this work uses a single latent space, meaning aspects
of variation within sets of hierarchically structured shapes
are entangled.  Recently, ~\cite{yang2020dsm} used separate VAEs for encoding the structure and the geometry of point clouds separately, meaning it is possible to condition output geometry on structure and vice versa. While this enables sampling conditioned on existing output characteristics to some extent, geometry and structure specific to all sub-trees of the output hierarchy are still entangled in the same latent space.  

In contrast, we learn a sequence of
individual probability distributions that correspond to each
LSD of hierarchically structured 3D shapes, allowing a subhierarchy
to be regenerated by sampling and decoding from some chosen level onward.

\subsection{Sequential latent space models} 

Generative models for data that is inherently sequential often couple each input term with a corresponding latent variable (e.g. ~\cite{goyal2017zforce}; applications to: image captioning in \cite{aneja2019sequential,desphande2019fast}; dialog generation in \cite{serban2017hierarchical}; handwriting in \cite{chung2015recurrent}). Such models can be applied to other kinds of data by imposing a sequence, including Laplacian pyramid levels from images ~\cite{denton2015deep}, sequences of resolutions ~\cite{wang2018pix2pixHD}, multi-scale feature representations ~\cite{vahdat2020NVAE}. 

Inspired by such methods, we model the generation process
of hierarchically structured 3D objects as a sequence of generations of geometric parts conditioned on their parents. Each input node is mapped to a latent variable according to its depth in the hierarchy. Each latent variable is responsible for the generation process at a given depth of the
output hierarchy.

\begin{figure}
\begin{center}
\includegraphics[width=1.0\linewidth]{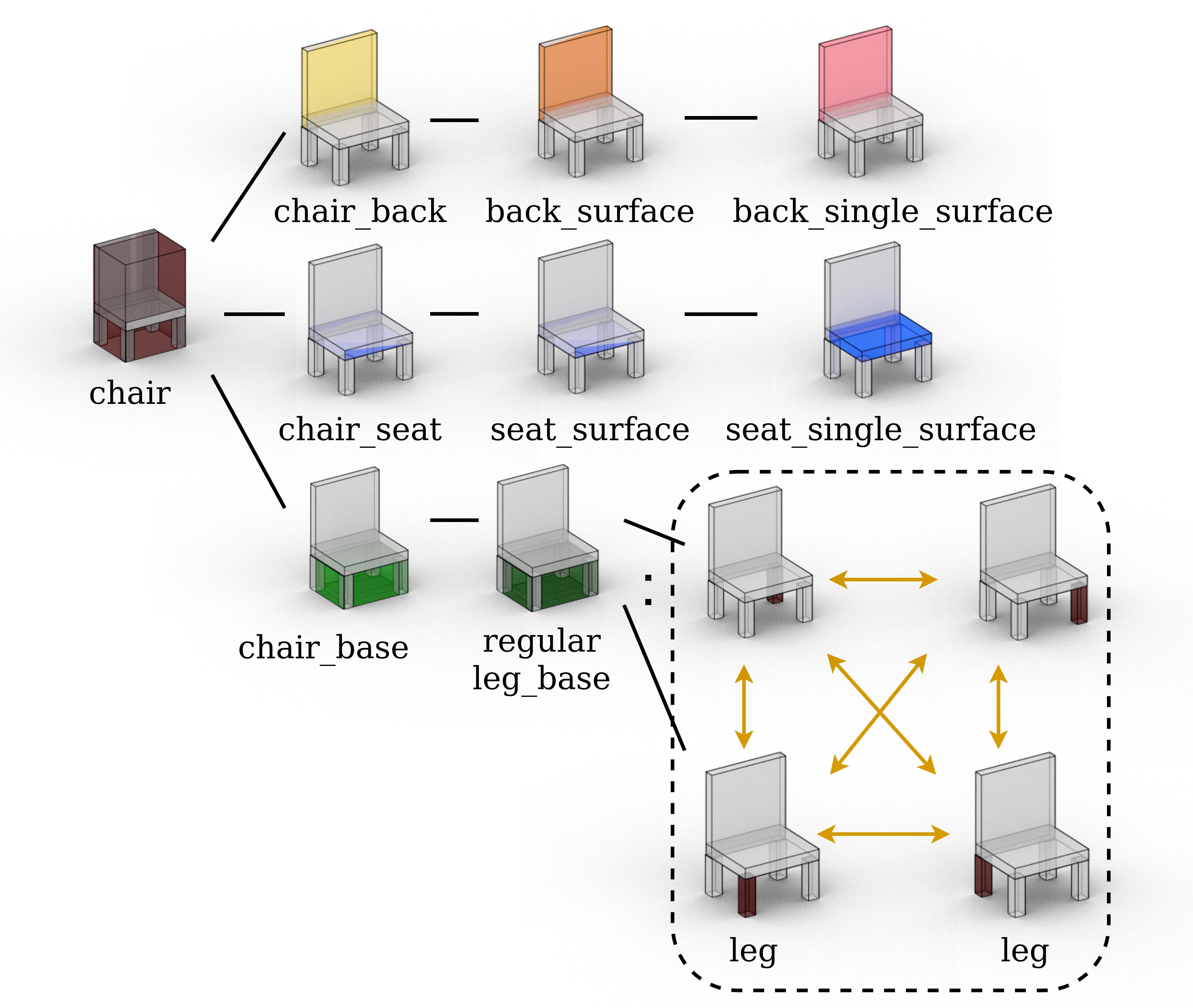}
\end{center}
   \caption{An example of a PartNet part hierarchy data sample. Parts in $\mathbf{P}$ with corresponding geometry (e.g. bounding boxes here) and semantic labels are connected with edges in $\mathbf{H}$ that form an n-ary tree structure (black lines) and edges in $\mathbf{R}$ (orange bi-directional arrows).}
\label{parthierarchy1} 
\end{figure} 

\begin{figure*}
\begin{center}
\includegraphics[width=1.0\linewidth]{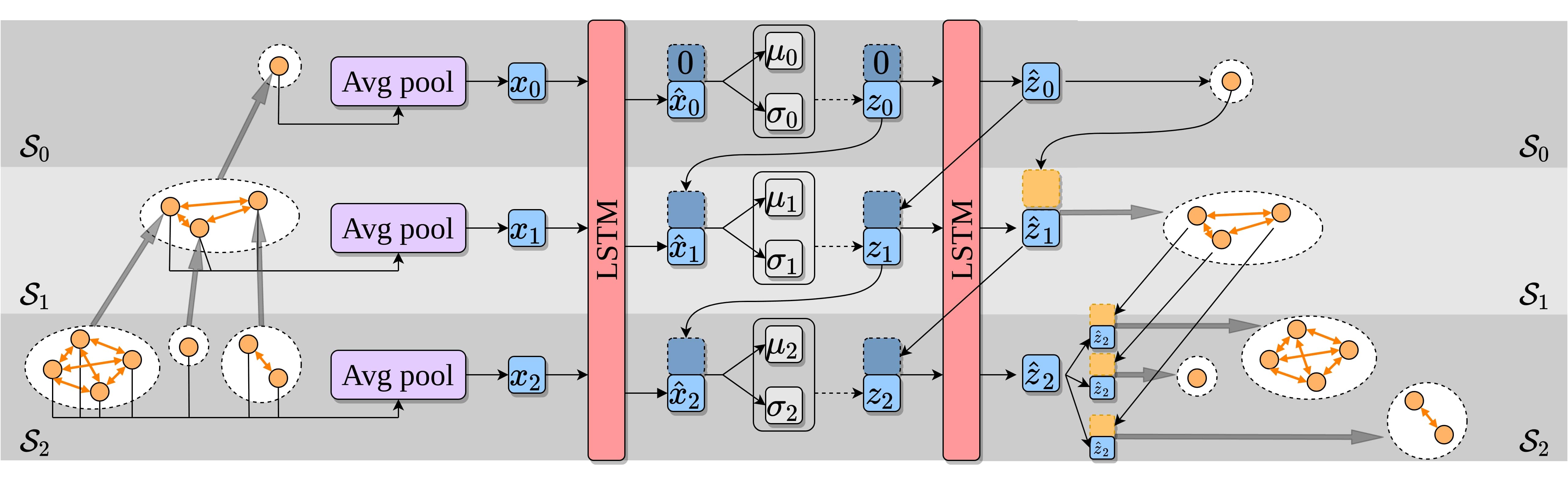}
\end{center}
   \caption{Left: We augment a StructureNet encoder by encoding an input hierarchy into a sequence of latent spaces. After an input shape is encoded into a hierarchy of feature vectors $\mathbf{F}$ via a graph encoder $g_{enc}$ (light gray arrows), geometry features of intermediary nodes are pooled according to their depth in the hierarchy and transformed into a sequence of latent codes sampled from depth-specific approximate posterior distributions. Right: we decode a sequence of latent codes by first putting these through an LSTM to obtain vectors that govern output geometry at each depth of the output hierarchy. These are concatenated with each node's geometry feature from the previous depth before putting them through an MLP $g_{dec}$ (light gray arrows) to obtain the geometry of their children.}
\label{methodfigure:short}
\end{figure*}

\section{Method}

\subsection{Shape structure} 

We retain the representation of 3D shapes used in StructureNet \cite{mo2019structurenet} that we visualize in Figure \ref{parthierarchy1}. Shapes are represented as a set of parts $\mathbf{P}$ and two sets of edges ($\mathbf{H}$,$\mathbf{R}$). $\mathbf{H}$ connects parts in an n-ary tree structure, whose root node corresponds to the shape as a whole and where each node's children correspond to the parts making up their parent. Edges in $\mathbf{R}$ connect children of a same parent and represent geometric relationships that can be of one of the following types: adjacency or rotational, translational or reflective symmetry. Each part is associated with a semantic label and with geometry in either the form of an oriented 3D bounding box or a 3D point cloud. Finally, the geometry associated with the leaf nodes in the hierarchy $\mathbf{H}$ constitutes the directly observable geometry of the object. 

\subsection{StructureNet}

StructureNet is a hierarchical graph network VAE that builds a generator for such shapes. An input is encoded into a latent vector via a bottom-up procedure: an MLP encodes the geometry of the leaf nodes of $\mathbf{H}$ into feature vectors. Each graph formed by the feature vectors and edges in $\mathbf{R}$ is encoded via a Graph Convolutional Network (GCN) $g_{enc}$, creating a feature vector associated with its parent in $\mathbf{H}$. This procedure is recurred following $\mathbf{H}$, creating a set of feature vectors $\mathbf{F}$ for which there is a one-to-one mapping with the set of parts $\mathbf{P}$. The final feature vector $f$ associated with the root node of $\mathbf{H}$ is used to obtain a latent vector $z$ as in standard VAE architectures, i.e. by sampling from $\mathcal{N}(\mu(f),\sigma(f))$ where $\mu(.)$ and $\sigma(.)$ are two MLPs.

Decoding a latent vector consists of recursively unpooling feature vectors into graphs via a recursive GCN decoder  $g_{dec}$ whose edges are part of the output $\mathbf{R}$. Individual MLPs are used to predict the geometry and semantic label for each feature vector; another MLP predicts whether a feature vector is associated with a leaf node in the output hierarchy $\mathbf{H}$. If it is predicted to be a leaf node, the recursion stops.

\subsection{LSD-StructureNet} 

While StructureNet and other methods can generate realistic and diverse part hierarchies, they cannot re-generate sub-hierarchies of a given output as the entirety of an output shape $\mathcal{S} = (\mathbf{P},\mathbf{H},\mathbf{R})$ is governed by one holistic generative process $p(\mathcal{S}|z)$. In other words, a given latent vector $z$ controls the entirety of an output $\mathcal{S}$ without there being any natural partitioning of $z$ that would enable modifying parts of the output without modifying the shape as a whole. 

We overcome this limitation by learning individual, per-depth generative processes for the geometry and structure $\mathcal{S}_d = (\mathbf{P}_d,\mathbf{H}_d,\mathbf{R}_d)$ of a shape at depth $d$ conditioned on that of the previous depth, i.e. $p(\mathcal{S}_d|z_d,\mathcal{S}_{d-1})$. As we show, this will allow us to re-generate sub-hierarchies. 

In the following, let $z_{<d} := (z_0,...z_{d-1})$. Following the formalism of ~\cite{aneja2019sequential}, we model the joint distribution $p_{\theta}(\mathcal{S})$ = $p_{\theta}((\mathcal{S}_d)_{d\geq0})$ with parameters $\theta$ and assume it factors into $\prod_{d\geq0}{p_{\theta}(\mathcal{S}_d|\mathcal{S}_{d-1})}$. We also assume multiple possible sub-hierarchies can be created for any object part at any given hierarchy depth, and model this possibility using a sequence of latent variables $\mathbf{z} = (z_d)_{d>0}$. We can thus decompose the joint via:
\begin{equation}
\begin{aligned}
    p_{\theta}(\mathcal{S}) = & \sum_\mathbf{z} p_{\theta}(\mathcal{S},\mathbf{z}) \\ 
                      = & \sum_\mathbf{z} \prod_d p_{\theta}(z_d|z_{<d},\mathcal{S}_{<d}) p_{\theta}(\mathcal{S}_d|\mathcal{S}_{<d},z_{\leq d}) 
\end{aligned}
\end{equation} 

As with standard VAEs, we approximate the posterior $p_{\theta}(z_d|z_{<d},\mathcal{S}_{<d})$ with an encoder $q_{\phi}(z_d|z_{<d},\mathcal{S}_{<d})$. We proceed to detail how we model the encoder, and the decoder  $p_{\theta}(\mathcal{S}_d|\mathcal{S}_{<d},z_{\leq d})$.

\textbf{Encoder.}  As with a StructureNet encoder, we encode parts $\mathbf{P}$ into feature vectors $\mathbf{F}$ recursively via $g_{enc}$. However StructureNet does not map inputs into a sequence of latent variables as we seek to do. As shown in Figure ~\ref{methodfigure:short}, we couple shapes $\mathcal{S}$ with per-depth latent vectors $z_d$ by firstly mapping each $\mathcal{S}_d$ into a geometry feature vector $x_d$. We do this by simply aggregating geometry features $\mathbf{F}$ at depth $d$ in part hierarchy $\mathbf{H}$ via an average-pooling layer.   

In order to model dependency between successive terms of the resulting sequence $\mathbf{x}$, we pass it through an LSTM, resulting in $ \mathbf{\hat{x}}$. We decompose the approximate posterior $q_{\phi}(z_d|z_{<d},\mathcal{S}_{<d})$ into:
\begin{equation}
q_{\phi}\big(z_d|z_{<d},\mathcal{S}_{<d}\big) = \prod_{i\leq d} q_{\phi}\big(z_i|z_{i-1},\mathcal{S}_{i-1}\big)
\end{equation} 
where each $q_{\phi}(z_i|z_{i-1},\mathcal{S}_{i-1})$ is parameterized at each depth $i\leq d$ via:
 \begin{equation}
     q_{\phi}\big(z_i|z_{i-1},S_{i-1}\big) =  \mathcal{N}\Big(\mu_i\big([\hat{x}_i,z_{i-1}]\big),\sigma_i\big([\hat{x}_i,z_{i-1}]\big)\Big)
 \end{equation} 
where $\mu_i(.),\sigma_i(.)$ are MLPs and $[,]$ signifies the concatenation operation. The encoding of $\mathcal{S}_d$ into the sequence of latent spaces is thus obtained by 
 \begin{equation}
     z_d \sim q_{\phi}\big(z_d|z_{d-1},\mathcal{S}_{d-1}\big) 
 \end{equation}
\textbf{Decoder}. A StructureNet decoder recursively unpools feature vectors $f_{i,d} \in \mathbf{F}_d$ indexed by $i$ at depth $d$ in $\mathbf{H}$ via $g_{dec}$. As $g_{dec}$ is deterministic, it cannot be used to produce diverse sets of child graphs. In contrast, we use a probabilistic decoder $p_{\theta}(\mathcal{S}_d|\mathcal{S}_{<d},z_{\leq d})$ to obtain all child graphs at $d$ via $\mathbf{z}$. We assume $f_{i,d}$'s child graph $c(f_{i,d})$ is obtained independently of all other nodes of $\mathcal{S}$ at depth $d$ given $z_d$ i.e. 
\begin{equation} \label{inddec}
    p_{\theta}\big(\mathcal{S}_d|\mathcal{S}_{<d},z_{\leq d}\big) = \prod_i p_{\theta}\big(c(f_{i,d})|f_{i,d},z_{\leq d}\big)
\end{equation}.  

To decode $\mathbf{z}$, we first put it through a decoding LSTM to obtain $\mathbf{\hat{z}} = LSTM(z_0,([z_d,\hat{z}_{d-1}])_{d\geq1})$. We then obtain the output hierarchy recursively, modelling $p_{\theta}(c(f_{i,d})|f_{i,d},z_{\leq d})$ via
\begin{equation} \label{decodeg}
    c(f_{i,d})= g_{dec} \big([f_{i,d},\hat{z}_{d}] \big) 
\end{equation} 

As with StructureNet, semantic labels and bounding box or point cloud geometry are predicted from feature vectors using MLPs, and a further MLP is used to predict whether a feature vector corresponds to a leaf node in $\mathbf{H}$, in which case recursion terminates. 

\textbf{Learning.} We train our encoder and decoder end-to-end, by encoding an input shape $\mathcal{S}$ into a sequence of latent vectors $\mathbf{z}$ which is decoded back into a shape. We choose a prior $p(\mathbf{z})$ that factors over hierarchy depths, i.e. $p(\mathbf{z}) = \prod_{d\geq0}{p(z_d|z_{<d},\mathcal{S}_{<d})}$ and model each ${p(z_d|z_{<d},\mathcal{S}_{<d})}$ as a unit Gaussian distribution $\mathcal{N}(0,1)$. As with a standard VAE, we minimize the variational regularization loss $\mathcal{L}_{var}$ that minimizes the distance between the approximate posterior and the prior by coercing each of the approximate posterior's components to be close to a unit Gaussian, i.e.
\begin{equation} 
    \sum_{d \geq 0} \mathcal{D}\bigg(\mathcal{N}\Big(\mu_d \big([\hat{x}_d,z_{d-1}]\big),\sigma_d \big([\hat{x}_d,z_{d-1}]\big)\Big) , \mathcal{N}(0,1)\bigg) 
\end{equation}
where $\mathcal{D}$ signifies KL-divergence. The entire network is trained with the sum of the variational regularization loss and the standard StructureNet reconstruction and structure consistency losses:
\begin{equation}
    \mathcal{L} = \mathcal{L}_{var} +  \mathcal{L}_{recon} + \mathcal{L}_{sc}
\end{equation}

See the supplementary material for the details of $\mathcal{L}_{recon}$,  $\mathcal{L}_{sc}$ and other implementation details. 

\begin{figure}[t]
\begin{center}
\includegraphics[width=1.0\linewidth]{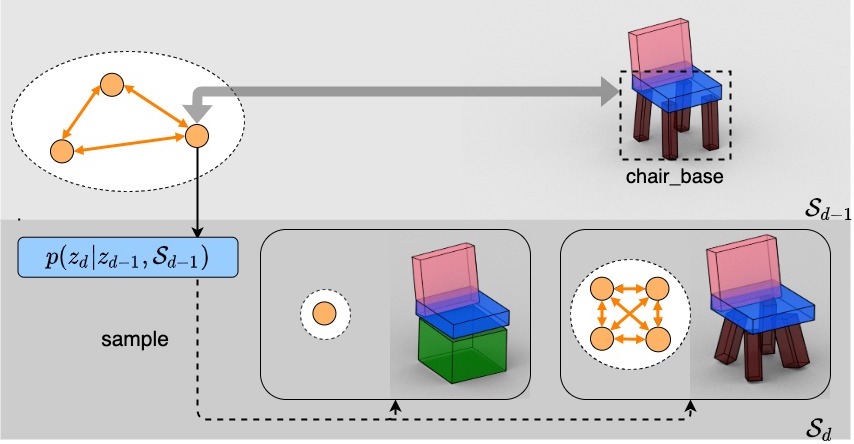}
\end{center}
   \caption{Multiple possible subhierarchies for any given node in a hierarchy can be generated by sampling from each per-LSD latent space below the node's and generating a graph of child nodes.}
\label{resampling}
\end{figure}

\textbf{Inference.} At test-time, we successively sample each term of the sequence of latent vectors $\mathbf{z}$  from the prior $\{p(z_d|z_{d-1},\mathcal{S}_{d-1})\}_{d\geq 0}$  which is then decoded to produce an output shape $\mathcal{S}$. Note that regenerating any sub-hierarchy of $\mathcal{S}$ whose root node is $f_{i,d}$ is as simple as re-sampling $z_{\geq d}$ and recursively applying Equation ~\ref{decodeg} to generate its children.

\section{Experiments}

We can use our method to generate sets of hierarchies that differ only from an existing ``conditioning'' hierarchy with regards to sub-hierarchies of our choosing. These sets should be plausible and diverse across a broad range of conditioning hierarchies and sub-hierarchies, and easily obtainable. Comparison with existing generative models for part hierarchies can not be made directly as they do not have this capability, though standard generative models are capable of indirectly conditioning new outputs on existing ones via rejection sampling, i.e. by sampling and decoding until an output whose sub-hierarchy is sufficiently ``close'' to an existing one is found.  

Following previous work, we train the network on the Chair, Table and Storage Furniture category training splits of PartNet, which is the only 3D shape dataset introduced thus far whose elements are organized according to hierarchies. We associate nodes in PartNet hierarchies with either bounding boxes or point clouds, and train individual networks for each.

\subsection{Generating and evaluating conditional samples}
\label{gencon}

\textbf{Conditional sampling.} Recall that we can conditionally sample an output hierarchy $\mathcal{S}$ by re-generating the hierarchy of a conditioning shape  $\mathcal{S}^{c}$ from an intermediary node $f_{i,D}$ at depth $D$. By applying this procedure across all $i$, we can re-generate the entirety of the hierarchy at depth $D$ and beyond, the sampled hierarchy being by design structurally, geometrically and semantically identical to the original hierarchy up until depth $D$, i.e.  $\mathcal{S}_{<D} = \mathcal{S}^{c}_{<D}$.   It is desirable for samples from these conditional distributions to be diverse across conditioning $\mathcal{S}_{<D}$ and depths $D$, as computer-aided design involves making choices at a wide range of structurally local levels.

Direct comparison with StructureNet in this regard is not possible as it does not model $p(\mathcal{S}|\mathcal{S}_{<D},z_{\leq D})$ and cannot perform such a conditioning sampling strategy.  To obtain sets of samples comparable to those obtainable through our method, we resort to an approximate conditional sampling strategy, retrieving hierarchies that are as similar as possible to the conditioning hierarchy via rejection sampling.  This strategy will yield outputs in non-constant time in any case and likely more slowly than via our method. Of interest is whether these outputs are as plausible and diverse as those obtained via our method, and to what extent the severity of the rejection criteria impacts sampling time.

\begin{table*}
\begin{center}
\begin{small}
\setlength{\tabcolsep}{10pt}
\begin{tabular}{l|l|l|cc|cc|cc|cc}
\toprule
\multirow{2}{*}{Category} & \multirow{2}{*}{Method} & \multirow{2}{*}{\parbox{1.5cm}{Sampling time (s) $\downarrow$}} & \multicolumn{2}{c|}{FPD$\downarrow$} & \multicolumn{2}{c|}{Structural Div.$\uparrow$} & \multicolumn{2}{c|}{Geometric Div.$\uparrow$} & \multicolumn{2}{c}{Semantic Div.$\uparrow$}\\ 
~ & ~ & ~ & $D$=1 & $D$=2 & $D$=1 & $D$=2 & $D$=1 & $D$=2 & $D$=1 & $D$=2 \\
\midrule
\multirow{2}{*}{Chair}
& SNet~\cite{mo2019structurenet} & 0.67 & 41.7 & 34.8 & \textbf{73.5} & \textbf{76.0} & 52.6 & \textbf{54.5} & \textbf{50.3} & \textbf{54.0} \\
& LSD-SNet & \textbf{0.05} & \textbf{40.4} & \textbf{34.4} & 69.3 & 70.6 & \textbf{55.4} & 54.3 & 49.0 & 48.2\\  
\hline
\multirow{2}{*}{Table}
& SNet~\cite{mo2019structurenet} & 3.0 & 85.5 & 76.3 & 26.2 & 27.1 & 76.8 & 79.1 & 12.6 & 13.1\\
& LSD-SNet & \textbf{0.04} & \textbf{44.7} & \textbf{48.9} & \textbf{56.5} & \textbf{76.3} & \textbf{90.7} & \textbf{85.2} & \textbf{29.9} & \textbf{42.6}\\ 
\hline
\multirow{2}{*}{Storage}
& SNet~\cite{mo2019structurenet} & 3.5 & 66.6 & \textbf{79.2} & 51.5 & 55.1 & 61.3 & 63.2 & 23.1 & 25.4\\
& LSD-SNet & \textbf{0.05} & \textbf{56.9} & 88.1 & \textbf{78.3} & \textbf{88.8} & \textbf{63.6} & \textbf{68.4} & \textbf{44.7} & \textbf{63.5}\\ 
\bottomrule
\end{tabular}
\end{small}
\end{center}
\caption{Sampling time, realism and diversity of sampled $(\mathcal{S}^{c}_{i,j})_{i<1000,j<100}$ characterized by bounding box geometry and generated conditionally on $\mathcal{S}^{c}_{i}$ at depths $D=1$ and $D=2$. Sampling time is averaged over 40  $\mathcal{S}^{c}_{i}$ and 20  $\mathcal{S}^{c}_{i,j}$. $\downarrow$ signifies `lower is better'. }
\label{conditionalsampling}
\end{table*} 
 
\begin{figure*}
\begin{center}
\setlength{\tabcolsep}{0pt}
\begin{tabular}{cc}
\adjustbox{trim={.0\width} {.05\height} {.0\width} {.02\height},clip}{\includegraphics[width=0.45\linewidth]{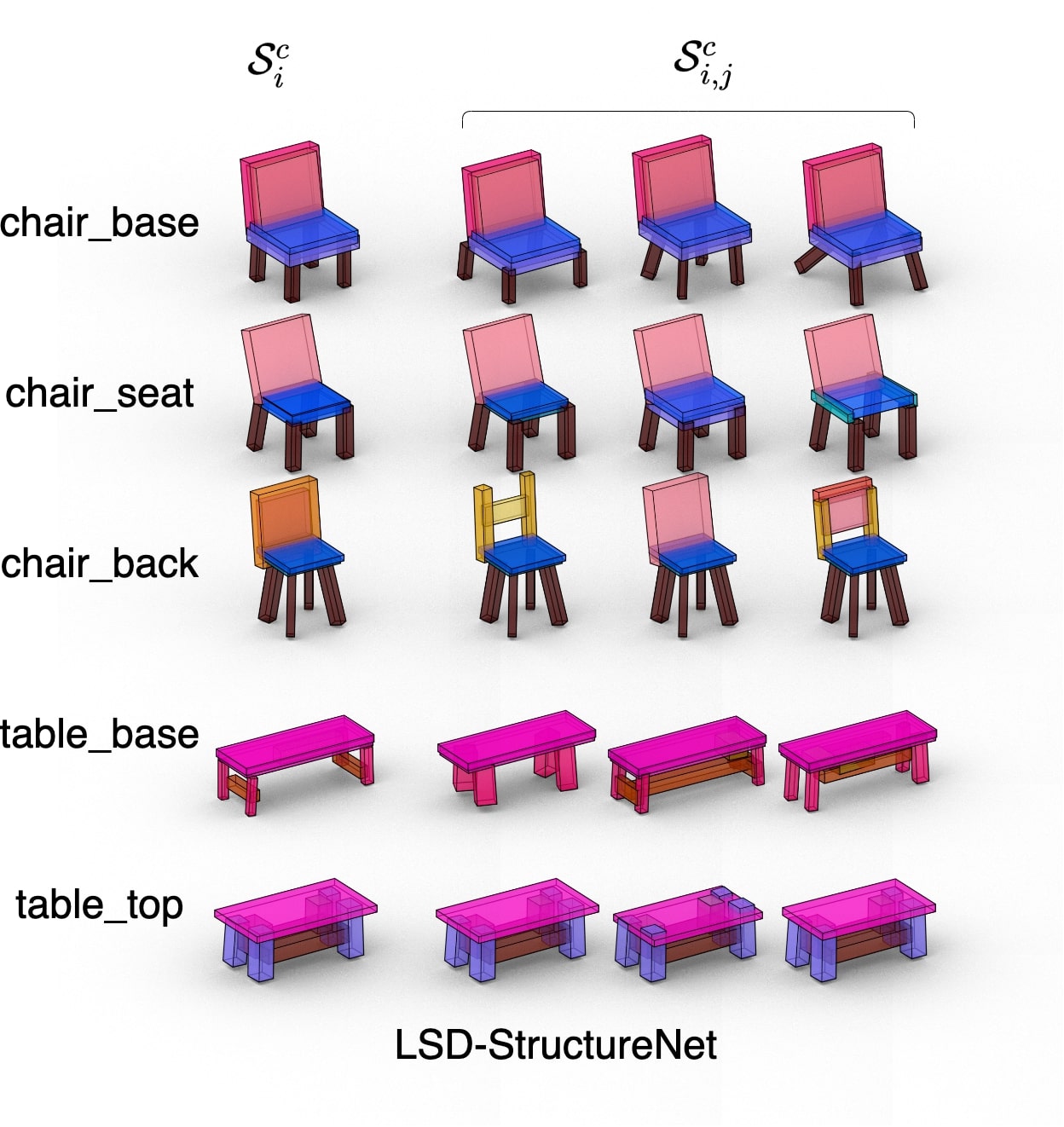}} & \adjustbox{trim={.0\width} {.05\height} {.0\width} {.02\height},clip}{\includegraphics[width=0.45\linewidth]{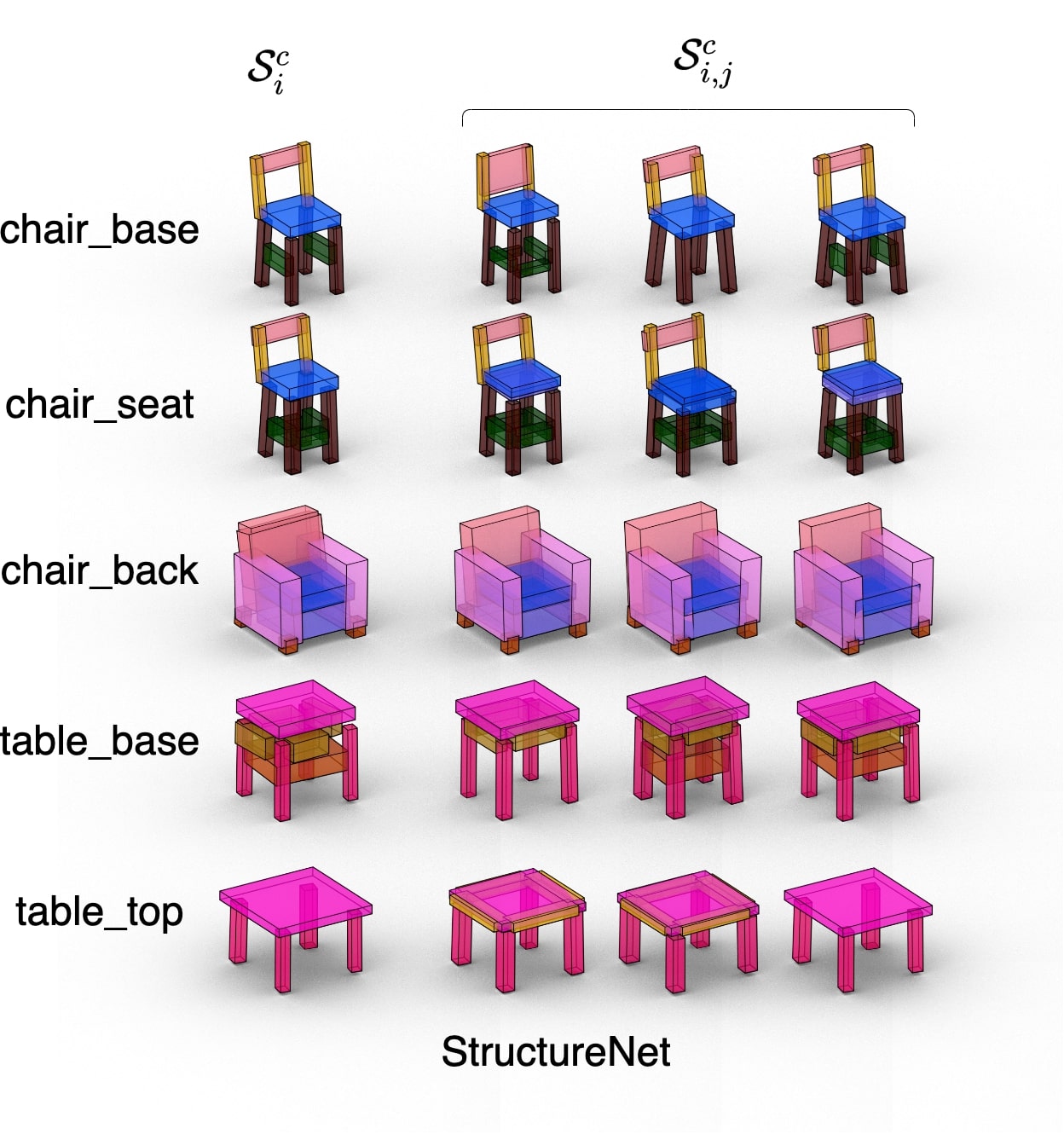}}
\end{tabular}
\end{center}
   \caption{LSD-StructureNet and StructureNet conditional samples whose structure is visualized in Figure \ref{parthierarchy1}. Given a conditioning $\mathcal{S}^{c}_i$ and semantic label corresponding to an intermediary (non-leaf) part, we re-generate $\mathcal{S}^{c}_{i,j}$ that only differ with regards to the part corresponding to the semantic label and its children in the hierarchy. StructureNet cannot achieve this, e.g. the structure of \texttt{chair\char`_back} varies noticeably among StructureNet conditional samples in the top row. LSD-StructureNet and StructureNet samples are obtained via the procedure visualized in Figure \ref{resampling} and rejection sampling respectively.  Our conditional samples are generally diverse and plausible at the level of the semantic label (rows 3 and 4) except when local, conditional diversity is not present in the data (row 5). Tables \ref{conditionalsampling} and \ref{resultshz} confirm that our samples are faithful to the data. While there is not a one-to-one correspondence between the $\mathcal{S}^{c}_i$ here, the lower variation between  StructureNet samples is likely the consequence of correlation, section \ref{gencon}. Our approach can perform conditional sampling more efficiently than StructureNet and in constant time (Figure \ref{plot}).}
\label{genconfig}
\end{figure*} 
 
\textbf{Conditional samples.}  We obtain a set of conditioning shapes at each depth $D$ via both methods by sampling 1000 shapes. Write  $\mathcal{S}^{c}_{i}$ the $i$th such conditioning shape. It is straightforward to obtain shapes conditioned on $\mathcal{S}^{c}_{i}$ from LSD-StructureNet by simply sampling 100 different $\mathcal{S}$ from $p(\mathcal{S}|\mathbf{z},\mathcal{S}^{c,i}_{>D})$ (write $\mathcal{S}^{c}_{i,j}$ the $j$th such conditionally generated shape), resulting in a set of $1000\times100$ output hierarchies $(\mathcal{S}^{c}_{i,j})_{i<1000,j<100}$. 

We obtain an analogous set from StructureNet via the following rejection sampling procedure: for each $\mathcal{S}^{c}_{i}$, sample $\mathcal{S}$ repeatedly from $p(\mathcal{S}|z)$, until 100 $\mathcal{S}$ are found that satisfy a) $\mathcal{S}_{<D}$ and $\mathcal{S}^{c,i}_{<D}$ are structurally and semantically identical (i.e. for both shapes, $\mathbf{H}_{<D}$, $\mathbf{R}_{<D}$ and the semantic labels of $\mathbf{P}_{<D}$  are all identical) and b) if the distances between the union of geometries associated with the nodes of $\mathcal{S}^{\prime}_{<D}$ and $\mathcal{S}_{<D}$ is below some tolerance $\epsilon$.

We sample latent vectors $z$ within distance $\eta$ of the latent vector used to produce each $\mathcal{S}^{c}_{i}$. We do this to increase the efficiency of our sampler so we have enough StructureNet samples to evaluate.  However, the samples must be correlated, and so biased to understate the variation of the conditional distribution. In this sampling procedure, $\epsilon$ and $\eta$ govern the trade-off between geometric fidelity of the conditioning shapes $\mathcal{S}^{c}_{i}$ and $\mathcal{S}^{c,i}_{<D}$ and the time it takes to sample $\mathcal{S}^{c,i}_{<D}$. 

\textbf{Metrics.} For a given conditioning shape $\mathcal{S}^{c}_{i}$, we evaluate diversity among its  $\mathcal{S}^{c}_{i,j}$ using 3 different metrics:

\begin{enumerate}
    \item \emph{structural} diversity, or the number of different hierarchical structures among the $\mathcal{S}^{c}_{i,j}$ 
    \item \emph{geometric} diversity, or the average chamfer distance between $\mathcal{S}^{c}_{i,j}$ i.e.
    
    \begin{equation}
        \frac{1}{100}\sum_{j,j^{\prime}<100} d_{\mathrm{chamfer}}(\mathcal{S}^{c}_{i,j},\mathcal{S}^{c}_{i,j^{\prime}})
    \end{equation} 
    
    \item \emph{semantic} diversity, or the number of different sets of semantic labels present among $\mathcal{S}^{c}_{i,j}$ 
    
\end{enumerate}

We also seek for the resulting set of $(\mathcal{S}^{c}_{i,j})_{i<1000,j<100}$ to be realistic, which we measure via FPD (Frechet Point Cloud Distance, \cite{shu20193d}). This metric calculates the distance between the set of generated samples and the test split of real-world hierarchies. The point cloud representations of the hierarchies in both sets are passed through a PointNet encoder, resulting in sets of feature encodings with means $\mu_s, \mu_t$ and covariances $\sigma_s, \sigma_t$. The FPD distance $d_{\mathrm{FPD}}$ is calculated via
\begin{equation}
    d_{\mathrm{FPD}} = \|\mu_s - \mu_t\|^2 + \mathrm{Tr}(\sigma_s + \sigma_t - 2\sigma_s\sigma_t)
\end{equation} 
where $\mathrm{Tr}$ signifies the trace in the context of matrices.

\textbf{Quantitative evaluation.} We report results for the Chair, Table and Storage Furniture PartNet categories in Table ~\ref{conditionalsampling}. As the vast majority of generated objects have hierarchies of depth 3 or less, we generate sets of $(\mathcal{S}^{c}_{i,j})_{i<1000,j<100}$ conditioned on depths $D = 1$ and $D=2$ of our conditioning shapes and whose parts are represented with oriented bounding boxes. We heuristically select $\epsilon = 0.05$ and $\eta=0.7$ as reducing these further meant that the rejection sampling procedure yielded too few shapes for evaluation purposes. We use a machine with a Tesla K40m GPU and an Intel Xeon E5-2680 CPU.

Results show that the conditional samples of LSD-StructureNet are generally more diverse than those of StructureNet while remaining realistic. Note however that as long as $\epsilon$ remains non-zero, the conditionally generated shapes will never be \emph{completely} faithful to the conditioning shapes up until depth $D$, biasing this evaluation in favor of StructureNet as its outputs exhibit differences between $\mathcal{S}^{c}_{i,j}$ up until depth $D$, increasing the diversity scores, which is particularly manifest for the Chair category.

\textbf{Qualitative evaluation.}  Given a conditioning shape, we use our method to produce several output hierarchies by re-generating the conditioning hierarchy solely from one intermediary node, defined by a given semantic label. As shown in Figure \ref{genconfig}, this procedure results in a set of output hierarchies that are generally structurally, geometrically and semantically diverse at the given locality.

We obtain comparable outputs from StructureNet by adapting the rejection sampling procedure to pertain uniquely to the geometry, semantics and structure of the nodes of the hierarchy that do not correspond to those of the given semantic label. As shown in Figure \ref{genconfig}, these outputs are not completely faithful to the conditioning shapes due to the tolerance $\epsilon$ necessary to obtain outputs in a timely manner. They also exhibit lower diversity. This is because the geometry and semantics of nodes at varying positions in the hierarchy are all entangled within StructureNet's holistic latent space, meaning the restriction imposed by the approximate conditioning carries over to the rest of the hierarchy.

\begin{figure}[t]
\begin{center}
\includegraphics[width=1.0\linewidth]{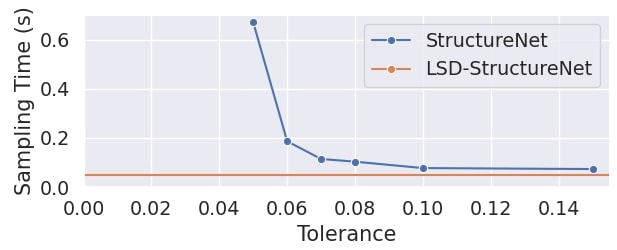}
\end{center}
   \caption{ Comparison of StructureNet and LSD-StructureNet  average  conditional sampling time, averaged across 20 Chair $\mathcal{S}^{c}_{i,j}$ and 40  $\mathcal{S}^{c}_i$ at depth $D=1$ as a function of tolerance $\epsilon$. Contrarily to StructureNet, LSD-StructureNet's sampling time is constant as it does not require rejection sampling.}
\label{plot}
\end{figure}  

\subsection{Speed of conditional sampling}

Here, we compare the constant time it takes to perform conditional sampling from LSD-StructureNet with the non-constant sampling time of StructureNet governed by tolerance $\epsilon$. With fixed $\eta=0.7$, we sample 20 Chair shapes conditionally from 10 different conditioning shapes and plot average sampling time as a function of $\epsilon$ in Figure ~\ref{plot}. StructureNet conditional sampling time rapidly increases when $\epsilon$ decreases. In contrast, LSD-StructureNet can easily perform conditional sampling without needing to reject samples, making its sampling time constant and far faster than StructureNet's.

\begin{table*}
\begin{center}
\begin{small}
\setlength{\tabcolsep}{14pt}
\begin{tabular}{ll|c|c|c|c|c|c}
\toprule
\multicolumn{2}{c|}{} & \multicolumn{3}{c|}{Bounding Box} & \multicolumn{3}{c}{Point Cloud} \\ 
\cline{1-8}
Category & Method & Coverage$\downarrow$ & Quality$\downarrow$ & FPD$\downarrow$ & Coverage$\downarrow$ & Quality$\downarrow$ & FPD$\downarrow$  \\
\midrule
\multirow{3}{*}{Chair}
& SNet~\cite{mo2019structurenet} & 26.0 & 47.4 & 57.0 & 24.6 & 44.6 & 53.1 \\
& LSD-SNet  & \textbf{25.4} & \textbf{47.2} & \textbf{39.3} & \textbf{24.0} & \textbf{43.3} & \textbf{51.4} \\    
\hline
\multirow{2}{*}{Table}
& SNet~\cite{mo2019structurenet} & 33.9 & 60.3 & 103.2 & 33.8 & \textbf{47.8} & 82.2 \\
& LSD-SNet  & \textbf{27.6} & \textbf{45.3} & \textbf{36.3} & \textbf{25.6} & 52.3 & \textbf{29.2} \\ 
\hline
\multirow{2}{*}{Storage}
& SNet~\cite{mo2019structurenet} & 4.4 & 58.0 & 71.6 & 4.4 & \textbf{55.0} & 86.2 \\
& LSD-SNet  & \textbf{4.4} & \textbf{51.4} & \textbf{63.9} & \textbf{4.1} & 55.7 & \textbf{60.9} \\
\bottomrule
\end{tabular}
\end{small}
\end{center}
\caption{Coverage, quality and FPD of a set of 1000 sampled shapes for each geometry type (bounding box and point cloud) from vanilla StructureNet (SNet) and our method (StructureNet augmented with LSD modelling capability). }
\label{resultshz}
\end{table*} 

\begin{figure*}
\begin{center}
\includegraphics[width=1.0\linewidth]{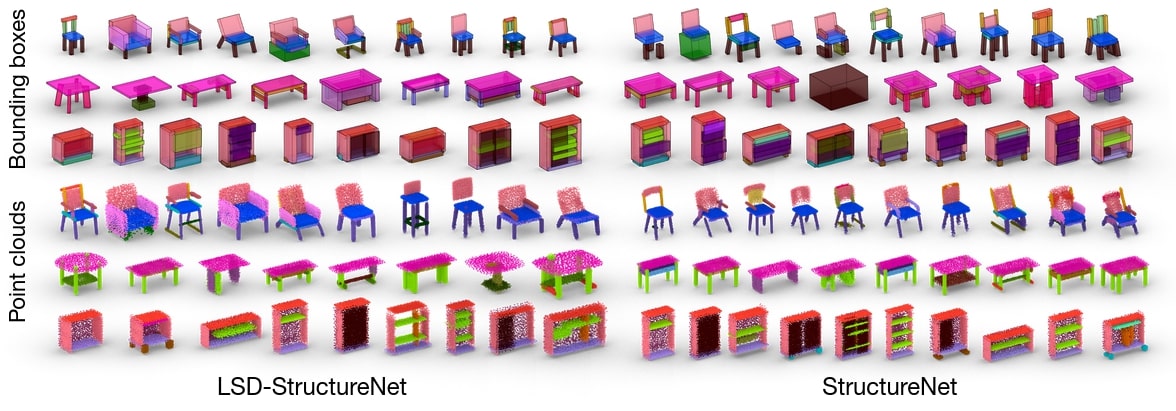}
\end{center}
   \caption{Chair, Table and Storage Furniture decoded samples from LSD-StructureNet (ours) and vanilla StructureNet (\cite{mo2019structurenet}). Our generated shapes are more diverse and plausible than those of StructureNet, reflecting the superior quantitative results of our method in Table \ref{resultshz}.}
\label{unconditional}
\end{figure*}

\subsection{Realism and diversity of samples generated without conditioning}

As attractive as the ability to rapidly and effectively perform conditional sampling may be, it should not
compromise the plausibility and diversity of sets of unconditional samples from the generative model. We show this is not an issue for LSD-StructureNet. 

\textbf{Metrics.} Following ~\cite{mo2019structurenet} we sample a set of 1,000 output shapes from the methods to be compared, which we term $G$. We compare this set to the test set $\mathcal{T}$ via

\begin{equation} 
     \mathrm{Coverage} := \sum_{\mathcal{S} \in \mathcal{T}} \min_{\mathcal{S}^{\prime} \in G} d(\mathcal{S}^{\prime},\mathcal{S}) 
\end{equation}

\begin{equation}
     \mathrm{Quality} := \sum_{\mathcal{S}^{\prime} \in G} \min_{\mathcal{S} \in \mathcal{T}} d (\mathcal{S}^{\prime},\mathcal{S}) 
\end{equation}

where $d$ is the chamfer distance between point cloud representations of $\mathcal{S}$. To obtain these when individual nodes in $\mathcal{S}$ are characterized by point clouds, we take the union of all point clouds of the leaf nodes of $\mathcal{S}$ and downsample via Furthest-Point Sampling (FPS) to reduce the number of points to 2048. When parts of $\mathcal{S}$ are characterized by bounding box geometry, we randomly sample 2048 points from the surfaces of the bounding boxes of the leaf nodes of $\mathcal{S}$. 

\textbf{Quantitative evaluation.} We display results in Table \ref{resultshz} and compare with StructureNet via quality, coverage and FPD for the test splits of categories Chair, Table and Storage Furniture . Our results show that LSD-StructureNet generates results that compare favorably to those of StructureNet. We report comparison to PQ-NET (~\cite{Wu_2020_CVPR}) in the Supplementary.

\textbf{Qualitative evaluation.} Figure \ref{unconditional} visualizes outputs across the 3 PartNet categories for both bounding box and point cloud modalities. Visually, our outputs are on par or better than StructureNet in terms of realism and diversity of the outputs.

\begin{table}
\begin{center}
\begin{small}
\setlength{\tabcolsep}{8pt}
\begin{tabular}{l|c|c} 
\toprule
Method & Coverage$\downarrow$ & Quality$\downarrow$ \\ 
\midrule
LSD$_0$ (SNet) & 26.0 & 47.4  \\ 
LSD$_1$ & 37.0 & 76.5  \\ 
LSD$_2$ & 28.5 & 52.2  \\
LSD$_3$ & 32.1 & 66.8  \\
\hline
LSD$_{\infty}$ (no-LSTM) & 73.5 & 102.5  \\ 

\hline
LSD$_{\infty}$ (LSD-SNet) & \textbf{25.4} & \textbf{47.2} \\ 
\bottomrule
\end{tabular}
\end{small}
\end{center}
\caption{We ablate our method by only modelling our probabilistic encoder $p(z_d|z_{<d},\mathcal{S}_{<d})$ and decoder $p(\mathcal{S}_d|\mathcal{S}_{<d},z_{\leq d})$ for $d\leq D_{max}$, decoding nodes at deeper depths with a deterministic StructureNet decoder (rows LSD$_{D_{max}}$), and by removing the encoding and decoding LSTMs (row no-LSTM) which we compare to our method LSD-SNet (bottom row).}
\label{ablation}
\end{table}

\subsection{Ablation study}

While these results show that outputs benefit holistically from modelling geometry and structure at intermediary depths of input hierarchies with individual latent spaces, it is not obvious that modelling \emph{all} hierarchy depths offers maximal benefits. We investigate this by training variations on our LSD-SNet architecture that only model hierarchy structure and geometry at \emph{some} intermediary  hierarchy depths. We do this by only coupling $\mathcal{S}_{\leq D_{max}}$ with corresponding latent variables $z_{<D_{max}}$ for varying $D_{max}$ . Here, our probabilistic decoder is only used to decode hierarchies up until depth $D_{max}$. Beyond this depth we continue recursive decoding with a StructureNet (deterministic) graph decoder $g_{dec}$. Note that $D_{max} = 0$ thus corresponds to vanilla StructureNet and $D_{max} = \infty$ corresponds to LSD-StructureNet. We further ablate our method by removing encoder and decoder LSTMs (i.e. $\mathbf{x} = \hat{\mathbf{x}}$ and $\mathbf{z} = \hat{\mathbf{z}}$). 

As shown in Table ~\ref{ablation}, our choice of components is justified by superior quality and coverage of LSD-StructureNet over its ablations. Rather than one decoder for the entire shape, for LSD$_1$-LSD$_3$ decoding is governed by a probabilistic decoder at $d \leq D_{max}$ and a separate deterministic one at $d \geq D_{max}$ . This means each decoder is more specialized, though each learns from less data, an empirically unfavorable trade-off. Furthermore, as illustrated in Figure \ref{methodfigure:short}, dependency of $z_i$ on $z_{i-1}$ is ensured by both a) the LSTM and b) $\mu_i$ and $\sigma_i$. The poor performance of LSD$_\infty$ without LSTM show that the LSTM is crucial for modeling the sequence of $z_i$.

\section{Conclusion}

This paper introduced LSD-StructureNet, an augmented version of StructureNet which is able to re-generate arbitrary parts of its 3D shape outputs without modifying the remainder of the shape. We have shown that LSD-StructureNet can generate shapes with diverse sets of re-generated parts that existing generative models cannot obtain without additional computational burden or modifying the conditioning shape. Furthermore, LSD-StructureNet achieves this capability while remaining on par with StructureNet in terms of the realism and diversity of its outputs. We hope that these results will incentivize the use of generative methods in 3D CAD design, which also involves creating variations on hierarchically structured 3D objects at a variety of positions in object hierarchies.

LSD-StructureNet inherits the limitations of StructureNet in that the outputs are not guaranteed to be geometrically or semantically valid and thus suffer from missing/duplicate parts and asymmetry. A potential solution which we leave for future work, could consist of integrating our approach with complementary generative approaches that explicitly model the fabrication process such as \cite{jones2020shapeAssembly}.

\textbf{Acknowledgements:} This work is supported in part by NSF Grant 2020227 (AI Institute:Planning: Construction). The authors thank Jyoti Aneja, Unnat Jain, Anand Bhattad, Alex Schwing and Aditya Sanghi for their input, comments and feedback on their work.

\newpage
\section{Supplementary Materials} 
\subsection{Implementation details} 

\subsubsection{StructureNet losses}

We train LSD-StructureNet with the loss
\begin{equation}
    \mathcal{L} = \mathcal{L}_{var} +  \mathcal{L}_{recon} + \mathcal{L}_{sc}
\end{equation}

where $\mathcal{L}_{var}$ is the variational loss defined in section 3.3. We proceed to briefly summarize the meanings of losses $\mathcal{L}_{recon}$ and $\mathcal{L}_{sc}$ that were originally introduced in ~\cite{mo2019structurenet}.

The reconstruction loss $\mathcal{L}_{recon}$ seeks to evaluate the best possible correspondence between input shape $\mathcal{S}$ and output $\mathcal{S}^{\prime}$. This is achieved by computing a linear assignment between parts of the input and output. Parts are matched via comparison of the geometries of the parts in each shape hierarchy. This assignment is then evaluated based on the geometry of the resulting pairs of parts, which are compared on the one hand via chamfer distance and via bounding box normals when the geometry is represented as bounding boxes. The existence of parts $\mathbf{P}$, edges $\mathbf{R}$, leaf nodes in $\mathbf{H}$ and semantic labels are also compared via cross-entropy.

The structure consistency loss $\mathcal{L}_{sc}$ seeks to enforce consistency between the geometric relationships $\mathbf{R}$ between siblings of a parent node in $\mathbf{H}$ (recall that these can be of types adjacency or rotational/translational/reflective symmetry) and the geometry $\mathbf{P}$ of output shapes $\mathcal{S}$. Symmetries are evaluated by computing an affine transformation between the bounding boxes fitted to the point clouds corresponding to the parts connected by each edge in $\mathbf{R}$, that transforms one part such that the symmetry is satisfied. The chamfer distance between the transformed and non-transformed point clouds are summed over $\mathbf{R}$ to constitute the final symmetry loss. For edges in $\mathbf{R}$ that are categorized as adjacencies, the loss is simply the minimum distance between geometries summed over $\mathbf{R}$. 
  
\subsubsection{Training details}

We encode shape part geometry (bounding boxes or point clouds) into 256-dim feature vectors. We dimension the encoding and decoding LSTMs such that they respectively output and input 512-dim feature vectors respectively. Aside from the input dimensions of the graph decoder $g_{dec}$ being 512 instead of 256, it and graph encoder $g_{enc}$ are identical to those of StructureNet. In terms of hyperparameters, we train LSD-StructureNet with the same weights attributed to each of the losses described in the previous paragraph, optimizer, learning rate, weight and learning rate decay, etc. as vanilla StructureNet. 
 
\begin{table}
\begin{center}
\begin{small}
\setlength{\tabcolsep}{7pt}
\begin{tabular}{l|c|c|c|c}
\toprule
Category & Method & Coverage$\downarrow$ & Quality$\downarrow$ & FPD$\downarrow$  \\ 
\midrule
\multirow{2}{*}{Chair}
& PQ-Net ~\cite{mo2019structurenet} & \textbf{8.9} & 116.5 & \textbf{28.9} \\
   & LSD-SNet & 25.4 & \textbf{47.2} & 39.3 \\    
\hline
\multirow{2}{*}{Lamp}
& PQ-Net ~\cite{mo2019structurenet} & 7.10 & 110.3 & \textbf{51.2} \\
& LSD-SNet & \textbf{5.9} & \textbf{67.8} & 141.3 \\  

\hline 
\end{tabular}
\end{small}
\end{center}
\caption{Quality, coverage and FPD of a set of 1000 sampled shapes characterized by bounding box geometry sampled from StructureNet, LSD-StructureNet and PQ-Net.}
\label{resultshz}
\end{table} 
 
\begin{figure}[t]
\begin{center}
\includegraphics[width=1.0\linewidth]{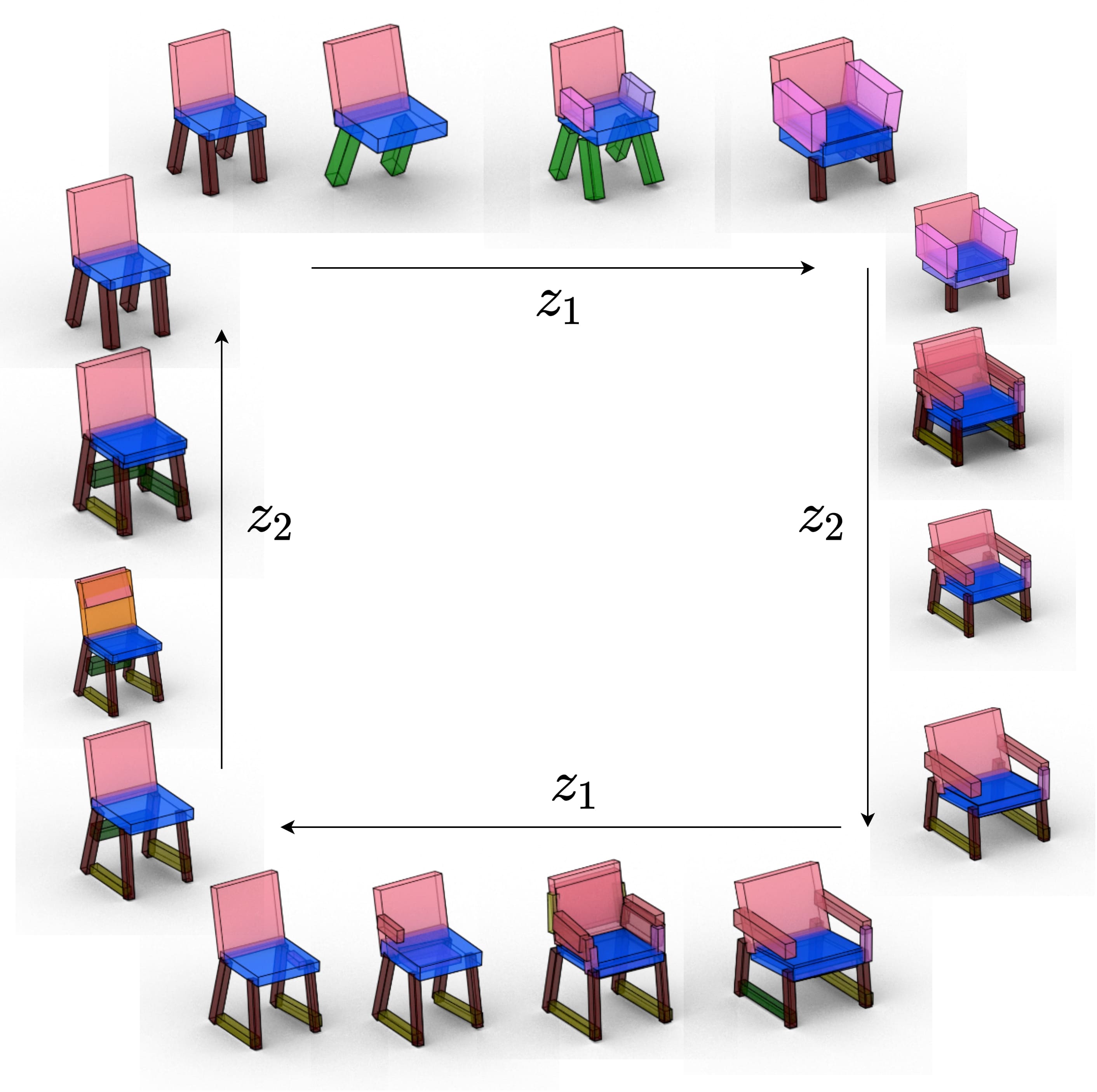}
\end{center}
   \caption{From a starting point at the top-left of the square decoded from a sequence of latent vectors $\mathbf{z}$, we interpolate between both $z_1$ and another latent vector $z_1^{\prime}$, and $z_2$ and another latent vector $z_2^{\prime}$ and decode the resulting sequences. }
\label{interp}
\end{figure}  

\begin{figure*}
\begin{center}
\includegraphics[width=1.0\linewidth]{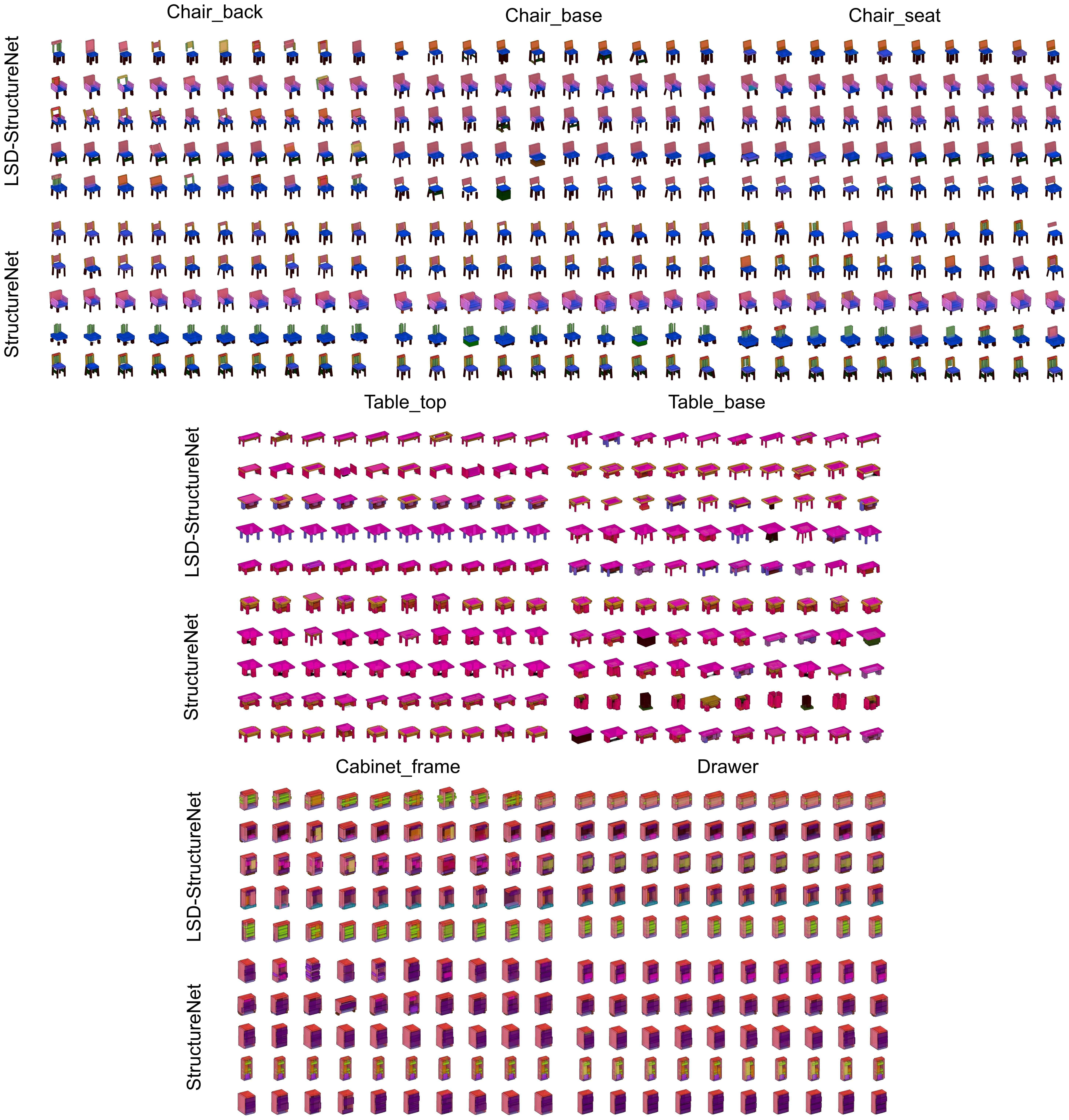} 
\end{center}
   \caption{Conditional outputs. For 5 shapes each for StructureNet and LSD-StructureNet (1st columns), we generate 9 other conditioning shapes (other columns) that only differ with respect to the subhierarchies of a given node (LSD-StructureNet) or are as similar as rejection sampling allows (StructureNet). We provide results for each penultimate node of Chair, Table and Cabinet/Storage hierarchies. Best viewed zoomed in.}
\label{cond}
\end{figure*}

\begin{figure*}
\begin{center}
\includegraphics[width=0.6\linewidth]{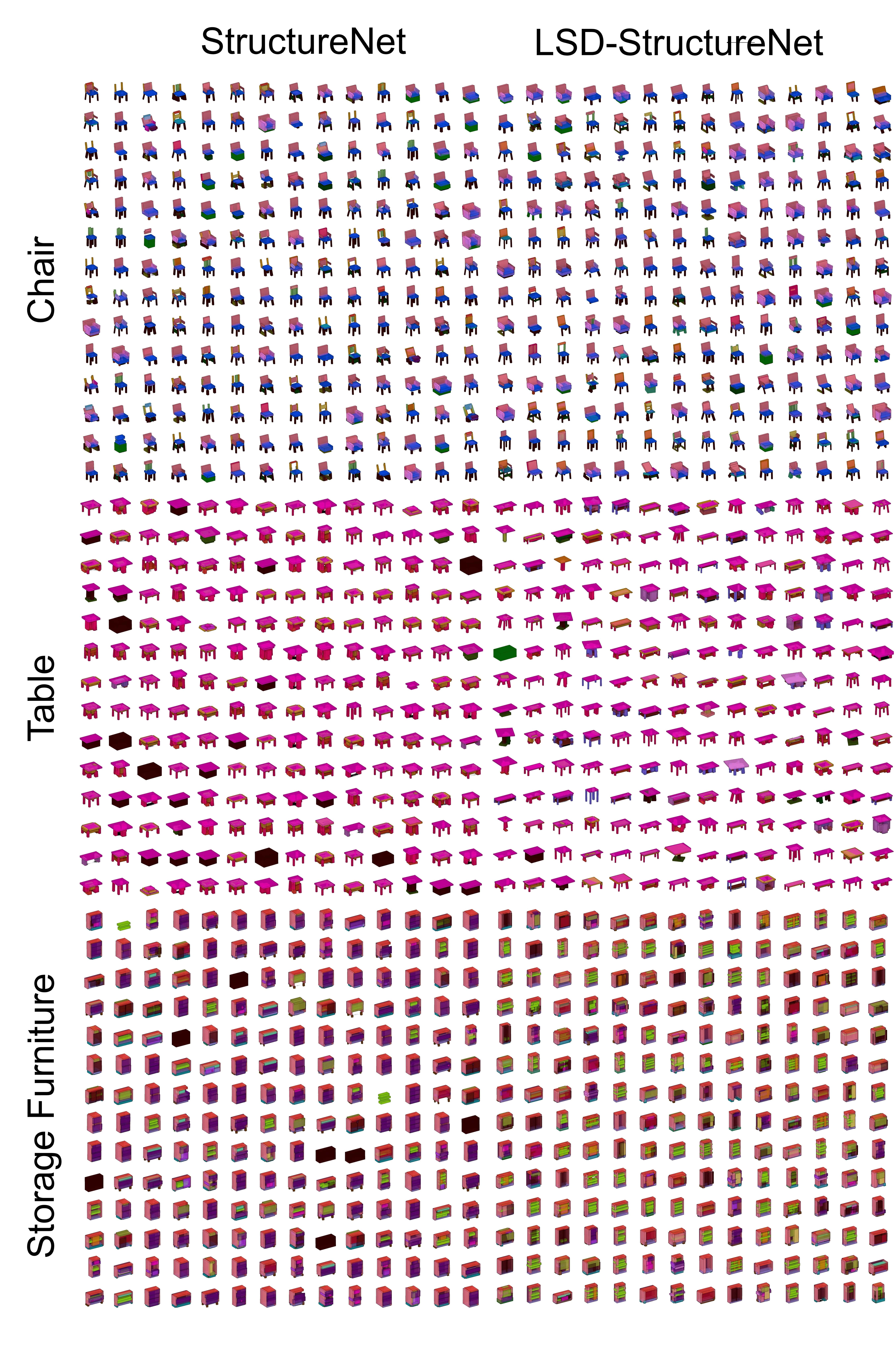}
\end{center}
   \caption{Unconditional outputs. We generate 100 Chairs, Tables and Cabinet/Storage shapes for both LSD-StructureNet and StructureNet. Best viewed zoomed in.}
\label{uncond}
\end{figure*}

\subsection{Comparison with PQ-Net} 
We supplement our comparison to vanilla StructureNet in Section ~\ref{resultshz} with a comparison to PQ-Net ~\cite{Wu_2020_CVPR}, a Seq2Seq model that can encode and decode sequences of directly observable part geometries of PartNet shapes (i.e. parts $\in \textbf{P}$ at leaf nodes of part hierarchies $\textbf{H}$ of shapes $\mathcal{S} = (\textbf{P}, \textbf{H},\textbf{R})$ ). We compare by sampling and decoding 1000 shapes from each method and report quality, coverage and FPD for shapes in the Chair and Lamp categories as these were the only two categories with available pretrained models for PQ-Net.

PQ-Net generates and is trained on sequences of parts as opposed to hierarchies. As such, it is not possible to augment PQ-Net directly so it can model intermediary levels of structural detail as we do with LSD-StructureNet. Despite this, while we outperform PQ-Net in terms of quality and coverage, it exhibits far stronger FPD than LSD-StructureNet, that mirror their similarly strong performance against StructureNet on similar metrics~\cite{mo2019structurenet}. This incentivizes potential future work consisting of consolidating the design choices of PQ-Net (obtention of latent space via Latent GAN instead of VAE and prediction of sequences, as opposed to part graphs) while retaining the hierarchical structure of LSD-StructureNet inputs and outputs.
 
\subsection{Visualizing outputs} 
We decode several different $\mathbf{z}$, linearly interpolating the 1st and 2nd vectors in the sequence between 2 extremes and visualizing the resulting outputs in Figure ~\ref{interp} to provide intuition as to the significance of the different latent spaces. Note that the parts of PartNet object hierarchies with semantic category \emph{chair arm} fade in (top row) or out (bottom row) when varying $z_1$, as they are situated at depth 1 of PartNet object hierarchies. The children of \emph{chair arm} parts are in this case leaf nodes with corresponding semantic categories \emph{arm sofa style} (top-right corner) and \emph{arm horizontal bar} (bottom-right corner) at depth 2, which is why varying $z_2$ produces interpolations between these two types (right column). In contrast, $z_2$ does not modify the structure at depth 1 and thus does not affect the presence of arms when varying it. We also qualitatively compare our outputs and those of StructureNet in Figures \ref{cond} and \ref{uncond}.

{\small
\bibliographystyle{ieee_fullname}
\bibliography{main}

\begin{thebibliography}{10}\itemsep=-1pt

\bibitem{aneja2019sequential}
Jyoti Aneja, Harsh Agrawal, Dhruv Batra, and Alexander Schwing.
\newblock Sequential latent spaces for modeling the intention during diverse
  image captioning.
\newblock In {\em ICCV}, 2019.

\bibitem{chung2015recurrent}
Junyoung Chung, Kyle Kastner, Laurent Dinh, Kratarth Goel, Aaron~C Courville,
  and Yoshua Bengio.
\newblock A recurrent latent variable model for sequential data.
\newblock In {\em NIPS}, 2015.

\bibitem{denton2015deep}
Emily~L Denton, Soumith Chintala, Rob Fergus, et~al.
\newblock Deep generative image models using a laplacian pyramid of adversarial
  networks.
\newblock In {\em NIPS}, 2015.

\bibitem{desphande2019fast}
Aditya Deshpande, Jyoti Aneja, Liwei Wang, Alexander~G Schwing, and David
  Forsyth.
\newblock Fast, diverse and accurate image captioning guided by part-of-speech.
\newblock In {\em CVPR}, 2019.

\bibitem{dubrovina2019composite}
Anastasia Dubrovina, Fei Xia, Panos Achlioptas, Mira Shalah, Raphael Groscot,
  and Leonidas Guibas.
\newblock Composite shape modeling via latent space factorization.
\newblock In {\em ICCV}, 2019.

\bibitem{gadelha2020learning}
Matheus Gadelha, Giorgio Gori, Duygu Ceylan, Radomir Mech, Nathan Carr, Tamy
  Boubekeur, Rui Wang, and Subhransu Maji.
\newblock Learning generative models of shape handles.
\newblock In {\em CVPR}, 2020.

\bibitem{gao2020tmnet}
Lin Gao, Tong Wu, Yu-Jie Yuan, Ming-Xian Lin, Yu-Kun Lai, and Hao Zhang.
\newblock {TM-NET}: Deep generative networks for textured meshes.
\newblock {\em arXiv:2010.06217}, 2020.

\bibitem{goyal2017zforce}
Anirudh Goyal Alias~Parth Goyal, Alessandro Sordoni, Marc-Alexandre
  C{\^o}t{\'e}, Nan~Rosemary Ke, and Yoshua Bengio.
\newblock Z-forcing: Training stochastic recurrent networks.
\newblock In {\em NIPS}, 2017.

\bibitem{groueix2018papier}
Thibault Groueix, Matthew Fisher, Vladimir~G Kim, Bryan~C Russell, and Mathieu
  Aubry.
\newblock A papier-m{\^a}ch{\'e} approach to learning 3d surface generation.
\newblock In {\em CVPR}, 2018.

\bibitem{jones2020shapeAssembly}
R.~Kenny Jones, Theresa Barton, Xianghao Xu, Kai Wang, Ellen Jiang, Paul
  Guerrero, Niloy~J. Mitra, and Daniel Ritchie.
\newblock {ShapeAssembly}: Learning to generate programs for 3d shape structure
  synthesis.
\newblock {\em SIGGRAPH Asia}, 39(6):Article 234, 2020.

\bibitem{kalogerakis2012probabilistic}
Evangelos Kalogerakis, Siddhartha Chaudhuri, Daphne Koller, and Vladlen Koltun.
\newblock A probabilistic model for component-based shape synthesis.
\newblock {\em TOG}, 31(4):1--11, 2012.

\bibitem{li17grass}
Jun Li, Kai Xu, Siddhartha Chaudhuri, Ersin Yumer, Hao Zhang, and Leonidas
  Guibas.
\newblock {GRASS}: Generative recursive autoencoders for shape structures.
\newblock {\em SIGGRAPH}, 36(4):1--14, 2017.

\bibitem{mo2019structurenet}
Kaichun Mo, Paul Guerrero, Li Yi, Hao Su, Peter Wonka, Niloy~J Mitra, and
  Leonidas Guibas.
\newblock Structurenet: hierarchical graph networks for 3d shape generation.
\newblock {\em TOG}, 38(6):1--19, 2019.

\bibitem{mo2020structedit}
Kaichun Mo, Paul Guerrero, Li Yi, Hao Su, Peter Wonka, Niloy~J Mitra, and
  Leonidas Guibas.
\newblock {StructEdit}: Learning structural shape variations.
\newblock In {\em CVPR}, 2020.

\bibitem{mo2019PartNet}
Kaichun Mo, Shilin Zhu, Angel~X. Chang, Li Yi, Subarna Tripathi, Leonidas
  Guibas, and Hao Su.
\newblock {PartNet}: A large-scale benchmark for fine-grained and hierarchical
  part-level {3D} object understanding.
\newblock In {\em CVPR}, 2019.

\bibitem{nash2020polygen}
Charlie Nash, Yaroslav Ganin, S.~M.~Ali Eslami, and Peter~W. Battaglia.
\newblock Polygen: An autoregressive generative model of 3d meshes.
\newblock {\em ICML}, 2020.

\bibitem{nash2017shape}
Charlie Nash and Christopher~KI Williams.
\newblock The shape variational autoencoder: A deep generative model of
  part-segmented 3d objects.
\newblock In {\em Computer Graphics Forum}, volume~36, pages 1--12, 2017.

\bibitem{park2019deepsdf}
Jeong~Joon Park, Peter Florence, Julian Straub, Richard Newcombe, and Steven
  Lovegrove.
\newblock Deepsdf: Learning continuous signed distance functions for shape
  representation.
\newblock In {\em CVPR}, 2019.

\bibitem{qi2017pointnet}
Charles~R Qi, Hao Su, Kaichun Mo, and Leonidas Guibas.
\newblock {Pointnet}: Deep learning on point sets for 3d classification and
  segmentation.
\newblock In {\em CVPR}, 2017.

\bibitem{schor2019CompoNet}
Nadav Schor, Oren Katzir, Hao Zhang, and Daniel Cohen-Or.
\newblock {CompoNet}: Learning to generate the unseen by part synthesis and
  composition.
\newblock In {\em ICCV}, 2019.

\bibitem{serban2017hierarchical}
Iulian~Vlad Serban, Alessandro Sordoni, Ryan Lowe, Laurent Charlin, Joelle
  Pineau, Aaron Courville, and Yoshua Bengio.
\newblock A hierarchical latent variable encoder-decoder model for generating
  dialogues.
\newblock In {\em AAAI}, 2017.

\bibitem{shu20193d}
Dong~Wook Shu, Sung~Woo Park, and Junseok Kwon.
\newblock {3D} point cloud generative adversarial network based on tree
  structured graph convolutions.
\newblock In {\em ICCV}, 2019.

\bibitem{socher2011parsing}
Richard Socher, Cliff Chiung-Yu Lin, Andrew~Y Ng, and Christopher~D Manning.
\newblock Parsing natural scenes and natural language with recursive neural
  networks.
\newblock In {\em ICML}, 2011.

\bibitem{tatarchenko2017octree}
Maxim Tatarchenko, Alexey Dosovitskiy, and Thomas Brox.
\newblock Octree generating networks: Efficient convolutional architectures for
  high-resolution 3d outputs.
\newblock In {\em ICCV}, 2017.

\bibitem{tian2018learning}
Yonglong Tian, Andrew Luo, Xingyuan Sun, Kevin Ellis, William~T. Freeman,
  Joshua~B. Tenenbaum, and Jiajun Wu.
\newblock Learning to infer and execute 3d shape programs.
\newblock In {\em ICLR}, 2019.

\bibitem{vahdat2020NVAE}
Arash Vahdat and Jan Kautz.
\newblock {NVAE}: A deep hierarchical variational autoencoder.
\newblock In {\em NeurIPS}, 2020.

\bibitem{valsesia2018learning}
Diego Valsesia, Giulia Fracastoro, and Enrico Magli.
\newblock Learning localized generative models for 3d point clouds via graph
  convolution.
\newblock In {\em ICLR}, 2018.

\bibitem{wang2018pix2pixHD}
Ting-Chun Wang, Ming-Yu Liu, Jun-Yan Zhu, Andrew Tao, Jan Kautz, and Bryan
  Catanzaro.
\newblock High-resolution image synthesis and semantic manipulation with
  conditional gans.
\newblock In {\em CVPR}, 2018.

\bibitem{wu2016learning}
Jiajun Wu, Chengkai Zhang, Tianfan Xue, Bill Freeman, and Josh Tenenbaum.
\newblock Learning a probabilistic latent space of object shapes via 3d
  generative-adversarial modeling.
\newblock In {\em NIPS}, 2016.

\bibitem{Wu_2020_CVPR}
Rundi Wu, Yixin Zhuang, Kai Xu, Hao Zhang, and Baoquan Chen.
\newblock Pq-net: A generative part seq2seq network for 3d shapes.
\newblock In {\em IEEE/CVF Conference on Computer Vision and Pattern
  Recognition (CVPR)}, June 2020.

\bibitem{wu2019sagnet}
Zhijie Wu, Xiang Wang, Di Lin, Dani Lischinski, Daniel Cohen-Or, and Hui Huang.
\newblock {SAGNet}: Structure-aware generative network for 3d-shape modeling.
\newblock {\em TOG}, 38(4):1--14, 2019.

\bibitem{yang2020dsm}
Jie Yang, Kaichun Mo, Yu-Kun Lai, Leonidas Guibas, and Lin Gao.
\newblock {DSM-Net}: Disentangled structured mesh net for controllable
  generation of fine geometry.
\newblock {\em arXiv:2008.05440}, 2020.

\bibitem{zhu2018scores}
Chenyang Zhu, Kai Xu, Siddhartha Chaudhuri, Renjiao Yi, and Hao Zhang.
\newblock {SCORES}: Shape composition with recursive substructure priors.
\newblock {\em TOG}, 37(6):1--14, 2018.

\end{thebibliography}
}

\end{document}